\documentclass[letterpaper, 10 pt, conference]{ieeeconf}

\IEEEoverridecommandlockouts                              

\usepackage{amsmath,amssymb,amsfonts}
\usepackage{bm}
\usepackage{tabularx}
\usepackage{graphicx}
\usepackage{textcomp}
\usepackage{xcolor}
\usepackage{color,soul}
\usepackage{subcaption}
\usepackage[ font=footnotesize]{caption}
\usepackage{url}
\usepackage{booktabs}
\usepackage{lipsum}
\usepackage{gensymb}
\usepackage{algorithm}
\usepackage{algpseudocode}
\usepackage{multicol}
\usepackage{amssymb}
\usepackage{dirtytalk}

\let\emptyset\varnothing

\algnewcommand{\algorithmicforeach}{\textbf{for each}}
\algdef{S}[FOR]{ForEach}[1]{\algorithmicforeach\ #1\ \algorithmicdo}

\usepackage{enumerate}
\usepackage{cite}
\usepackage{hyperref}
\hypersetup{linkbordercolor=green}
\usepackage[english]{babel}

\usepackage{nicefrac}
\usepackage{flushend}
\usepackage{textcomp}
\title{\LARGE \bf
Probabilistic Risk Assessment for Chance-Constrained Collision Avoidance in Uncertain Dynamic Environments
}

\author{Khaled A. Mustafa, Oscar de Groot, Xinwei Wang, Jens Kober, and Javier Alonso-Mora
\thanks{ The authors are with the Dept. of Cognitive Robotics, TU Delft, 2628 CD Delft, The Netherlands. {\tt\small Email:  k.a.mustafa@tudelft.nl}.
}
\thanks{This research was supported by funding from the Dutch Research Council NWO-NWA, within the “Acting under uncertainty” (ACT) project (Grant No. NWA.1292.19.298), and the European Union's Horizon 2020 research and innovation program within SAFE-UP project under Grant agreement 861570.}
}
\begin{document}

\maketitle
\thispagestyle{empty}
\pagestyle{empty}

\begin{abstract}
Balancing safety and efficiency when planning in crowded scenarios with uncertain dynamics is challenging where it is imperative to accomplish the robot's mission without incurring any safety violations. Typically, chance constraints are incorporated into the planning problem to provide probabilistic safety guarantees by imposing an upper bound on the collision probability of the planned trajectory. Yet, this results in overly conservative behavior on the grounds that 
the gap between the obtained risk and the specified upper limit is not explicitly restricted. To address this issue, we propose a real-time capable approach to quantify the risk associated with planned trajectories obtained from multiple probabilistic planners, running in parallel, with different upper bounds of the acceptable risk level. Based on the evaluated risk, the least conservative plan is selected provided that its associated risk is below a specified threshold. In such a way, the proposed approach provides probabilistic safety guarantees by attaining a closer bound to the specified risk, while being applicable to generic uncertainties of moving obstacles. 
We demonstrate the efficiency of our proposed approach, by improving the performance of a state-of-the-art probabilistic planner, in simulations and experiments using a mobile robot in an environment shared with humans.

\end{abstract}

\section{Introduction}

Mobile robots are appealed to work in complex environments shared with humans, such as smart warehouses \cite{amazon}, autonomous driving \cite{SD} and maritime transportation \cite{MI}. In these applications, the robot needs to progress toward its goal while safely avoiding static and dynamic obstacles. This task poses great challenges due to the fact that the robot needs to account for the possible uncertainties associated with the future predicted states of moving obstacles, as well as localization errors. These uncertainties make it difficult to decide whether the planned trajectories by the robot are safe or if given specifications, such as safety distance, are not violated. As a consequence, uncertain scenarios require mobile robots to find a reasonable trade-off between safety and efficiency. This gives rise to the
the problem of risk-aware motion planning in uncertain dynamic environments \cite{risk-aware, GP, Risk-Maps, Pek, analyticrisk, STL}.

In this paper, we address the problem of estimating the risk associated with the collision probability of mobile robots surrounded by moving obstacles and integrating the estimated risk in a local motion planning framework to plan collision-free trajectories while balancing risk and progress. In particular, the probability of collision is estimated by integrating over the spatial domain at which the robot's plan and obstacles' predicted states overlap. The proposed risk metric is, consequently, measured by the maximum risk value over different time instants within a prediction horizon. To that end, we integrate this risk metric into a probabilistic motion planning framework to enhance its efficiency, in terms of traveling time, while maintaining the estimated risk below a specified upper level. 
\subsection{Related Work}
\subsubsection{Collision Avoidance Under Uncertainty}
Optimization-based motion planning algorithms can plan collision-free trajectories in uncertain environments by incorporating the uncertain behavior of dynamic obstacles as constraints into the optimization problem. These algorithms can be classified into two common approaches, namely \textit{robust optimization} \cite{robust-optimization} and \textit{stochastic optimization} \cite{stochastic-optimization}. Robust optimization approaches are able to provide safety guarantees by rigorously accounting for bounded sets of uncertainties, that is the probability density function of the uncertainty is non-zero over a bounded domain of the robot's workspace and is zero elsewhere. However, since robust optimization accounts for all possible realization of the uncertainty, its behavior is too conservative and may lead to infeasible solutions in crowded scenarios \cite{Freezing}. On the contrary, stochastic optimization allows for the violation of the constraints as long as the probability of this violation is below an acceptable upper bound, which is specified through \textit{chance constraints} \cite{caltech}, \cite{Hai-MAV}, \cite{CC}. In this work, we rely on a stochastic optimization approach. 
\subsubsection{Safety Assessment in Motion Planning} One of the key components in safety analysis for motion planners is the risk metric that quantifies the risk level. For risk-aware motion planning algorithms, this risk metric usually indicates the collision probability due to, among others, the uncertain behavior of dynamic obstacles or imprecise localization of the robot. In \cite{GP}, Gaussian process regression is employed to build a probabilistic model of the environment which is used to construct a risk-aware cost function. This cost function is then encoded into an optimal motion planning algorithm. \cite{Risk-Maps} builds spatiotemporal probabilistic risk maps. These maps indicate how risky a planned trajectory (computed by a rapidly-exploring random tree algorithm) will be, and are used to plan the best possible future behavior that maximizes utility while minimizing risk. A similar idea is used in \cite{Pek} to estimate the risk of violating a predefined safety specification and encode it into a sampling-based trajectory planner \cite{Werling} to plan minimal-risk trajectories. A drawback of these approaches, however, is the high computational cost due to the extensive trajectory generation as well as the bias in the trajectory selection. Related to our risk definition, \cite{analyticrisk} and \cite{Wang} propose an analytic approach to calculate the probability of spatial overlap for a ground vehicle with dynamic obstacles at discrete times. Along the same line as our approach, \cite{STL} proposes a framework, using signal temporal logic, which provides probabilistic safety guarantees that can be embedded in a receding horizon controller. However, their approach is restricted to safety constraints on random variables with unimodal distributions.
Differently from the aforementioned approaches, in this paper, we propose an approach to incorporate a posterior risk assessment for planned trajectories into a probabilistic motion planning framework that applies to general probability distributions. From \cite{Hai-MAV}, \cite{CC}, \cite{SH-MPC}, it is noted that the \textit{observed risk} of the planned trajectory is much lower than the upper bound of the \textit{specified risk} in the chance constraint problem. This conservatism can be attributed to the collision probability marginalization of the planned trajectory. That is, the collision probability at each planning step along the horizon is independent \cite{Patil}. 

\subsection{Contribution}
To alleviate the over-conservatism problem with probabilistic motion planners, we propose the following:
\begin{enumerate}[(i)]
    \item Multiple probabilistic planners run in parallel with different upper bounds of the specified risk. The set of planners should include the planner where the upper bound of the risk is the desired one, which is the most conservative planner.
    \item An online posterior risk assessment is provided to quantify the risk associated with each planned trajectory from the multiple planners.
    \item Deploy the control commands from the planner with the least conservative behavior as long as its associated risk is below the specified risk of the most conservative planner.
\end{enumerate} 
In this way, the proposed approach provides probabilistic safety guarantees while achieving a closer bound to the specified risk, resulting in more efficient and less conservative performance.

\begin{figure*}
        \centering
        \subcaptionbox{Predicted positions of the robot and pedestrian at stage $k$ are visualized in faded orange and green, respectively. One realization of the pedestrian's uncertainty spatially overlaps with the robot's planned trajectory.\label{fig1:a}}{\includegraphics[width=2.1in]{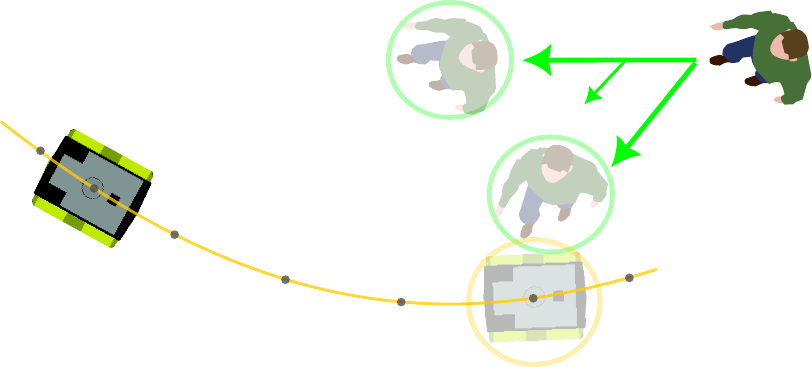}}\hspace{2.5em}%
        \subcaptionbox{Probability density function of the GMM representing pedestrian motion uncertainty at stage $k$. Only two modes are presented in this Fig.\label{fig1:b}}{\includegraphics[width=2.1in]{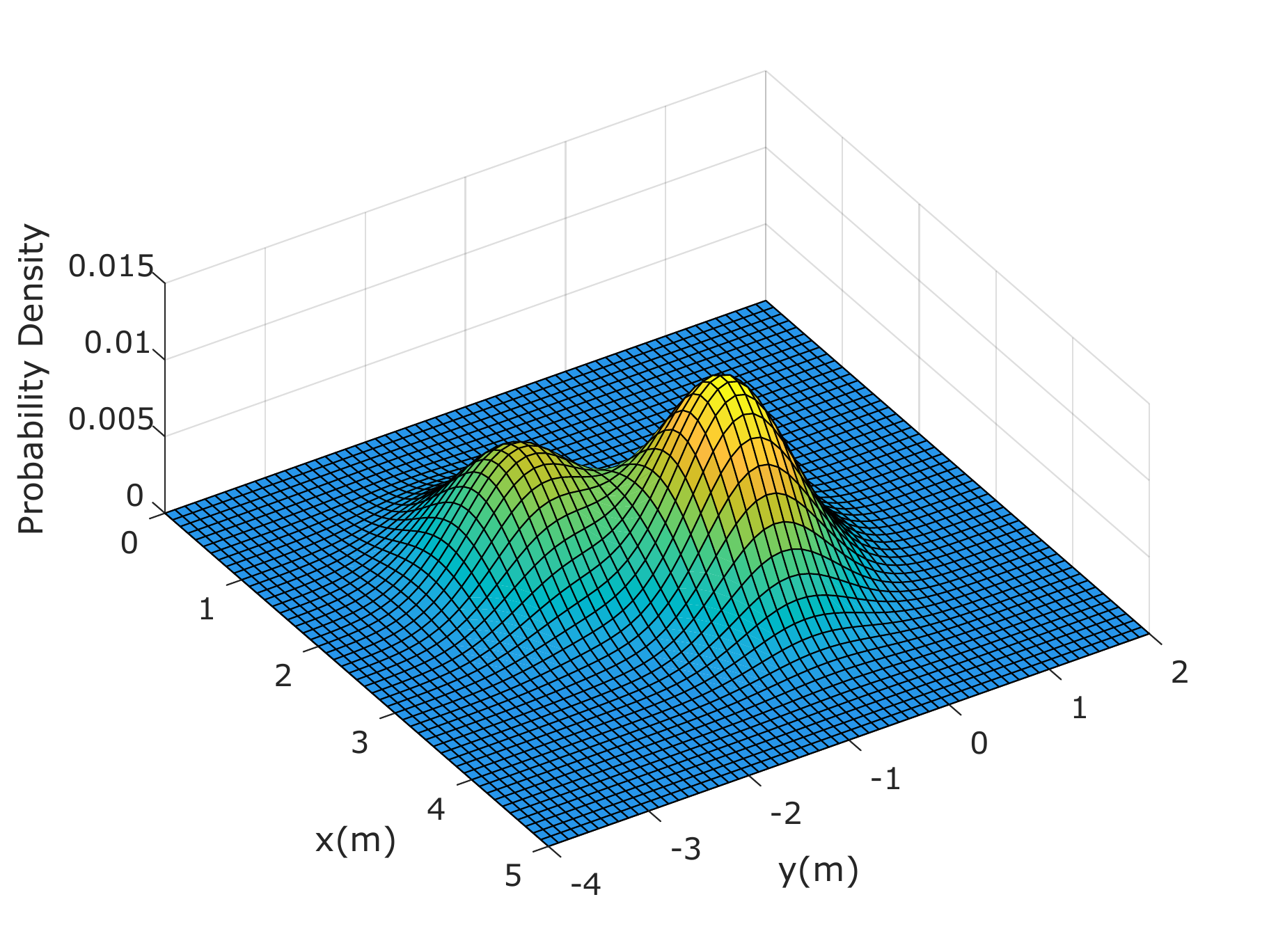}}
        \caption{An illustrating example of the proposed approach for one robot's disc, single pedestrian, and one realization of the associated uncertainty at stage $k$. As described in the results section, for GMM case, each pedestrian has $N+1$ modes where only two of them are visualized in Fig. \ref{fig1:a}, and Fig. \ref{fig1:b}. The estimated risk can be evaluated by integrating the PDF visualized in Fig. \ref{fig1:b} over a circular domain with radius $r = r_c + r_v$.} \label{proposed_approach}
\end{figure*}
\section{Preliminaries}
\label{preliminaries}
Throughout this paper, vectors, and matrices are expressed in bold, $\boldsymbol{x}$, and capital bold, $\boldsymbol{A}$, letters respectively. $||\boldsymbol{x}||$ is the Euclidean norm of $\boldsymbol{x}$, and the subscript $._k$ indicates the value at stage $k$.  
\subsection{Robot Model}
The dynamics of a ground robot moving in a 2D plane, $\mathcal{W} \in \mathbb{R}^2$, are modelled as a non-linear discrete-time system,
\begin{equation}
    \boldsymbol{x}_{k+1} = f(\boldsymbol{x}_k, \boldsymbol{u}_k),
\end{equation}
where $\boldsymbol{x}_k = \left[\boldsymbol{p}_k, \psi_k\right] \in \mathbb{R}^{n_x}$ and $\boldsymbol{u}_k \in \mathbb{R}^{n_u}$ denote the state and control input of the robot at stage $k$ respectively. The state of the robot $\boldsymbol{x}_k$ contains its position $\boldsymbol{p}_k = (x,y)$ and orientation $\psi_k$. 
The area occupied by the robot at state $\boldsymbol{x}_k$ is denoted by $\mathcal{O}(\boldsymbol{x}_k)$ which is approximated by the union of $n_c$ circles.  
\subsection{Dynamic Obstacle Model}
Each dynamic obstacle $v \in \mathcal{I}_v := \{1,...,n\}$ is represented by a circle with radius $r_v$. The probability measure associated with the uncertainty of the perception of the dynamic obstacles is denoted by $\mathbb{P}$ and defined over the probability space $\Delta$. 
Without loss of generality, the uncertainty associated with obstacle movement is modeled as a Gaussian Mixture Model,
\begin{equation}
    f_k^v(x,y) = \sum_{i=1}^{n}\phi_i f_{k,i}^v(x,y),
\end{equation}
where $n$ is the number of modes of the GMM, $\phi_i$ represents the weight of each mode such that $\sum_{i=1}^n \phi_i= 1$, and $f_{k,i}^v(.)$ is the probability density function of each mode with mean $\boldsymbol{\mu}_i$ and covariance $\boldsymbol{\Sigma}_i$.


\noindent \textbf{Assumption 1.} \textit{We assume that at each stage, a perception module provides the planner with a model of the probability.}
\subsection{Probabilistic Collision Avoidance}
\noindent \textbf{Definition 1. } (Chance-Constrained Collision Avoidance) \textit{Given a cost function $J$, the initial state of the robot $\boldsymbol{x}_0 = \boldsymbol{x}_\text{\textnormal{init}}$, and the state distribution of obstacles $v \in \mathcal{I}_v$, the objective is to compute optimal control inputs that guide the robot from its initial state to progress along a reference path, while the collision probability with the moving obstacles at each stage $k$ is below an acceptable threshold $\epsilon_k$. The resulting optimization problem is given by}
\begin{subequations}
\begin{alignat}{2}
\min_{\boldsymbol{u} \in \mathbb{U}} \quad & \sum_{k=0}^{N-1}J_k(\boldsymbol{x}_k, \boldsymbol{u}_k) + J_N(\boldsymbol{x}_N)\\
\textrm{s.t.} \quad & \boldsymbol{x}_0 = \boldsymbol{x}_{\text{init}},\\
 & \boldsymbol{x}_{k+1} = f(\boldsymbol{x}_k, \boldsymbol{u}_k), \quad \boldsymbol{x} \in \mathbb{X}, \boldsymbol{u} \in \mathbb{U}, \label{eq1b}\\
  &\mathbb{P} \left[||\boldsymbol{x}_k^d - \boldsymbol{\delta}_k^v||_2 > r, \forall d, v\right] \geq 1-\epsilon_k, \forall k, \label{CC}  
\end{alignat}
\end{subequations}
where $J_k(\boldsymbol{x}_k, \boldsymbol{u}_k)$ represents the stage cost of the robot, and $J_N(\boldsymbol{x}_N)$ denotes the terminal cost. States $\boldsymbol{x}_k$ and inputs $\boldsymbol{u}_k$ are bounded by the state and input constraint sets $\mathbb{X}$ and $\mathbb{U}$ respectively. $\boldsymbol{\delta}_k^v \in \Delta_k ^v$ is the realization of the uncertain position of obstacle $v$ at stage $k$. The radius $r$ is the summed radii for the robot's disc $d$, and obstacle $v$. The chance constraint, defined in \eqref{CC}, constrains the marginal probability of collision at each stage of the trajectory to be below the risk level $\epsilon_k$. In this paper, the stage cost $J_k(\boldsymbol{x}_k, \boldsymbol{u}_k)$ is defined by the Model Predictive Contouring Control framework proposed in \cite{bruno} to track a reference path, and a reference velocity while penalizing the control inputs. By solving the optimization problem, we obtain a locally optimal sequence of commands $\left[\boldsymbol{u}_k^*\right]_{k=0}^{k=N-1}$ to guide the robot along the reference path while
avoiding collisions with dynamic obstacles. Here it should be pointed out that a global reference path is assumed to be provided to our local planner by a means of a global planner. This reference path is composed of $M$ way-points $p_m^r = [x_m^r, y_m^r, \theta_m^r] \in \mathcal{W}$ with $m \in \{1,...,M\}$.

\section{Proposed Approach}
\label{proposed approach}
In this work, we aim to define a risk metric that can be incorporated into a probabilistic motion planner framework to balance safety and efficiency in a comprehensible way. This is motivated by the fact that state-of-the-art probabilistic planners, for navigation in environments with non-gaussian uncertainties, are overly conservative, e.g. \cite{CC}, \cite{SH-MPC}. In particular, we rely on scenario-based MPC proposed in \cite{SH-MPC} as our probabilistic planner to enhance its efficiency. Nevertheless, the proposed approach is agnostic to the deployed probabilistic planner and can be widely applicable. In scenario-based MPC, the risk bound $\epsilon_k$, 
at each stage $k$ is correlated to the number of samples drawn from the uncertainty. From \cite{SH-MPC}, it is noted that without manually tuning the number of samples extracted from dynamic obstacles uncertainty, the level of risk associated with the planned trajectory is much lower than the upper bound of the acceptable risk. This, in turn, results in conservative plans. Here it is worth pointing out that tuning the samples manually can only be done a posteriori and thus it is not suitable for online planning where the \textit{observed risk} is not known a priori. Therefore, we propose to quantify the risk associated with the planned trajectories from multiple scenario-based MPCs, running in parallel with different risk bounds, online and pick the least conservative plan as long as its associated risk is below a specified threshold. 
\subsection{Scenario-based MPC} 
Similar to \cite{SH-MPC}, since the chance constraint defined in \eqref{CC} is non-convex, we first linearize it with respect to the previously planned robot trajectory $\hat{\boldsymbol{x}}_k$. The linearization is applied locally at each stage $k$, and robot disc $d$. This results in 
\begin{subequations}
\begin{alignat}{2}
    &\boldsymbol{A}_k(\boldsymbol{\delta}_k, \hat{\boldsymbol{x}}_k) = \frac{\boldsymbol{\delta}_k - \hat{\boldsymbol{x}}_k}{||\boldsymbol{\delta}_k - \hat{\boldsymbol{x}}_k||},    b_k(\boldsymbol{\delta}_k, \hat{\boldsymbol{x}}_k) = \boldsymbol{A}_k^T\boldsymbol{\delta}_k - r,\\
    &\mathbb{P}\left[\boldsymbol{A}_k^T(\boldsymbol{\delta}_k, \hat{\boldsymbol{x}}_k)\boldsymbol{x}_k \leq b_k(\boldsymbol{\delta}_k, \hat{\boldsymbol{x}}_k)\right] \geq 1 - \epsilon_k, \forall k, \boldsymbol{\delta}_k \in \Delta_k, \label{LCC}
\end{alignat}
\end{subequations}
By linearizing the collision region with respect to $\hat{x}_k$, it can be seen that each scenario constraint in \eqref{LCC} defines a half-space. The free space of the scenario program is, in turn, formed by the intersection of these half-spaces resulting in a convex constraint, that spans a polytope $\mathcal{P}_k$, with respect to the robot's position. \\
\noindent Evaluating the chance constraints in a closed loop is not computationally feasible. Thus, it is aimed to formulate them into deterministic constraints using scenario optimization, resulting in a tractable constrained optimization problem that can be solved online in a receding horizon manner. As shown in \cite{SH-MPC}, the probabilistic chance constraints can be transformed to deterministic ones by leveraging a deterministic scenario program (SP) \cite{SP} for a finite set of samples/scenarios $\boldsymbol{\omega} = \left(\boldsymbol{\delta}^{(1)},...,\boldsymbol{\delta}^{(S)}\right)$, where each scenario is independently extracted from $\mathbb{P}$. Hence, the chance constraint problem can be reformulated as 
\begin{subequations}
\label{OCP}
\begin{alignat}{2}
\min_{\boldsymbol{u} \in \mathbb{U}} \quad & \sum_{k=0}^{N-1}{J_k(\boldsymbol{x}_k, \boldsymbol{u}_k) + J_N(\boldsymbol{x}_N)}\\
\textrm{s.t.} \quad & \boldsymbol{x}_0 = \boldsymbol{x}_{\text{init}},\\
& \boldsymbol{x}_{k+1} = f(\boldsymbol{x}_k, \boldsymbol{u}_k), \quad \boldsymbol{x} \in \mathbb{X}, \boldsymbol{u} \in \mathbb{U}, \label{eq1b}\\
  &\boldsymbol{A}_k^T(\boldsymbol{\delta}_k^i, \hat{\boldsymbol{x}}_k)x_k \leq b_k(\boldsymbol{\delta}_k^i, \hat{\boldsymbol{x}}_k), \forall k, i=1,..., \mathcal{S},  \label{SP}  
\end{alignat}
\end{subequations}
where the chance constraint \eqref{LCC} has been replaced with the deterministic constraints \eqref{SP} for each extracted scenario. The probability that the planned input $\boldsymbol{u}$ violates the predefined acceptable risk $\epsilon$ is defined as $V(\boldsymbol{u}^\ast)$ and upper bounded by a confidence level $\beta$.
This confidence bound is defined by
\begin{equation}
    \mathbb{P}^\text{S}\left[V(\boldsymbol{u}^\ast) > \epsilon(s) \right] \leq \sum_{s=0}^{S-1} \binom{S}{s} \left[1 - \epsilon(s)\right]^{S-s} = \beta, \label{V}
\end{equation}
where $\mathbb{P}^\text{S}$ is the product probability measure, given by $\mathbb{P}^\text{S} = \mathbb{P} \times\cdots\times\mathbb{P}$ (S times), and $s$ is the size of the \textit{support subsample}, that is the minimum number of samples that results in the same solution as the original sample $S$. In other words, if a scenario can be excluded from the scenario set $\boldsymbol{\omega}$ without affecting the optimizer solution, this scenario is then not part of the support subsample. \eqref{V} establishes a relationship between sample size, risk, and support subsample. The readers can refer to \cite{SH-MPC} for a comprehensive overview.

\subsection{Risk Assessment}
In this paper, the risk is defined as the probability of collision of each of the robot's discs with any of the moving obstacles.
\noindent Given the planned trajectory $\mathcal{T}$ of the robot for a controller, and the probability density function $f_k^v(x,y)$ that defines the uncertainty of the dynamic obstacle's movement in a 2D plane, it is possible to calculate the cumulative density function (CDF) for each obstacle $v$ at each stage $k$ along the prediction horizon by evaluating the integration of their associated probability density function at the robot's disc predicted position $\boldsymbol{x}_k^d$.\\
\noindent \textbf{Definition 2.} (Risk Metric) \textit{Let $ \mathcal{Z} $ denote the set of random variables representing the uncertainty of the pedestrians' motion in the} $x$ \textit{and} $y$ \textit{directions. The risk metric maps the distribution of the random variables to a real number indicating the probability of collision,} $\zeta: \mathcal{Z} \mapsto \mathbb{R}$, \textit{ by estimating their spatial overlap with the robot's plan $\mathcal{T}$. The probability of collision can, subsequently, be defined as}
\begin{equation}
    C_k^{v}(\boldsymbol{x}_k^d) =
    \iint_{x^d_k, y^d_k \in D}  f_k^{v}(x,y) dx dy, \forall{k}, v, d,
\end{equation}
This integration can be approximated numerically using the Monte Carlo method \cite{MC},
where the integration domain $D$ is defined as a circle whose center is located at the predicted vehicle pose $\boldsymbol{x}_k^d$ at stage $k$ along the prediction horizon, and its radius $r$ is the sum of the vehicle and obstacle radii.\\
After calculating the probability of collision for each pedestrian along the robot's planned trajectory, the predicted risk at the current time step is defined by maximizing the collision probability for all pedestrians at all planning horizon stages.
\begin{equation}
   \zeta = \max_{v \in \mathcal{I}_v, k, d} C_k^v (\boldsymbol{x}_k^d),  \label{max_risk}
\end{equation}

\noindent The max operator in \eqref{max_risk} ensures that the worst-case overlap is considered over the planned trajectory. The proposed approach is illustrated in Fig. \ref{proposed_approach}, for a single pedestrian and one realization of the associated uncertainty at stage $k$, and summarized in Algorithm \ref{alg:cap}, where $\emptyset$ indicates that no feasible solution is obtained from any of the controllers.

\begin{algorithm}
\caption{Risk-Aware scenario-based MPC}\label{alg:cap}
\begin{algorithmic}
\Require A set of scenario-based MPCs $\pi \in \mathcal{I}_\pi := \{1,...,n\}$ with different $\epsilon_i \in \{\epsilon_1,...,\epsilon_n\}$ values where they are defined in a descending order, and a predefined risk threshold $\epsilon_\circ = \epsilon_n$
\Ensure{ Control input command: $u = \emptyset$}
\While{$\boldsymbol{x}_0:=\boldsymbol{x}(t) \notin \mathcal{X}_\text{goal}$}
\State $\left[\boldsymbol{x}_k^\pi\right]_{k=0}^{k=N}, \left[\boldsymbol{u}_k^\pi\right]_{k=0}^{k=N-1} \gets \text{Solve } \eqref{OCP} \text{ simultaneously } \forall{\pi} $
\ForEach {$\pi \in \mathcal{I}_\pi$}
    
    \State Evaluate the estimated maximum risk $\zeta$ from \eqref{max_risk}
    \If {$\zeta < \epsilon_\circ$}
    \State $u \gets \boldsymbol{u}_0^\pi$
    \State \textbf{return}
    \EndIf
\EndFor
\If {$u = \emptyset$}
\State $u \gets \text{Deploy maximum deceleration}$
\EndIf
\EndWhile
\end{algorithmic}
\end{algorithm}


\section{Results}
\label{results}
 In this section, we describe our implementation of the proposed method, for a mobile robot navigating in a crowded environment shared with humans, and evaluate it in simulations and experiments.
 \subsection{Experimental Setup}
 \subsubsection{Software Setup}
 The motion planner is implemented as a ROS node in C++. Our simulations use the open-source ROS implementation of the Jackal Gazebo for the robot simulation. 
 To solve SP \eqref{OCP}, we use ForcesPro solver \cite{ForcesPro}. A horizon of $N = 20$ steps is defined, with a discretization step of 0.2 s, resulting in a time horizon of 4.0 s. The control rate is set to 20 Hz corresponding to a sampling time of 50 ms. The computer running the simulations is equipped with an Intel\textsuperscript{\textregistered} Core\textsuperscript{TM} i7 CPU@2.6GHz. The robot dynamics are described by a continuous-time second-order unicycle model \cite{model}.  The radius of each robot's circle is set to 0.325 m with $n_c = 2$, and the obstacle radius is set to 0.3 m.
 
 \begin{figure*}
\centering
\begin{subfigure}{.3\textwidth}
  \centering
  \includegraphics[width=0.95\linewidth]{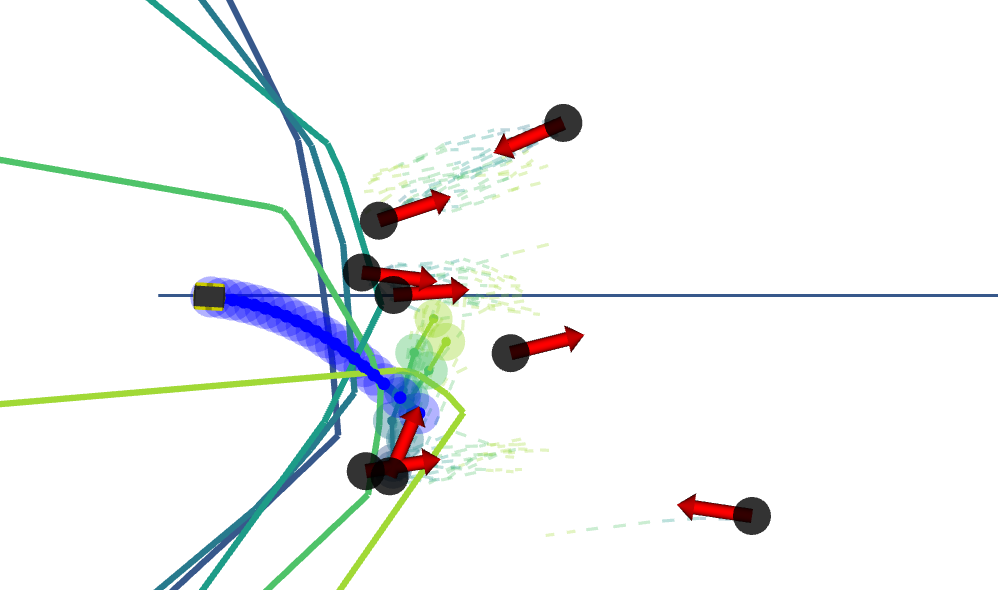}
  \caption{At $t=1.0$ s}
  \label{fig:sub1}
\end{subfigure}%
\begin{subfigure}{.3\textwidth}
  \centering
  \includegraphics[width=0.95\linewidth]{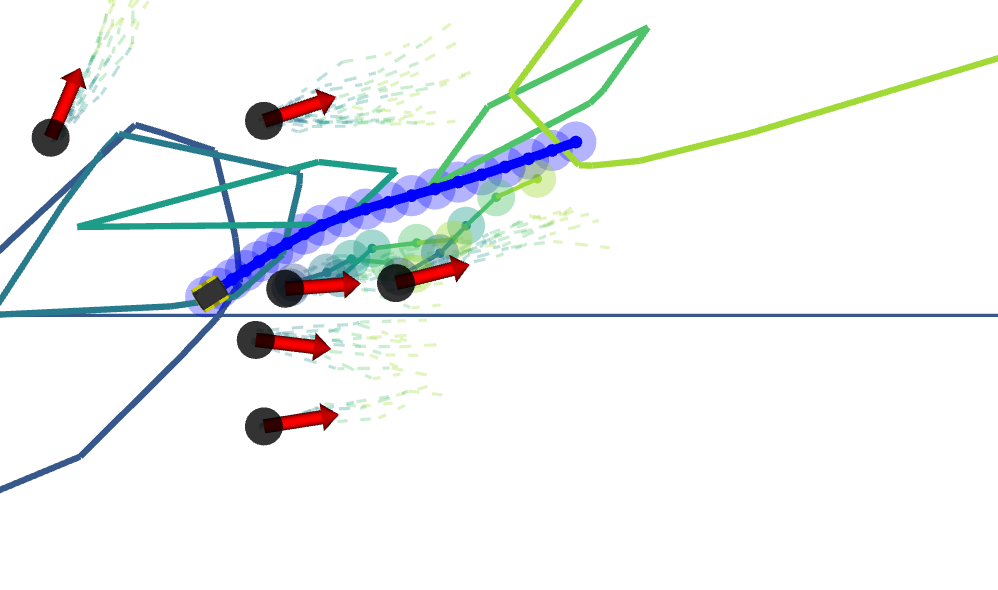}
  \caption{At $t=8.0$ s}
  \label{fig:sub2}
\end{subfigure}
\begin{subfigure}{.3\textwidth}
  \centering
  \includegraphics[width=0.95\linewidth]{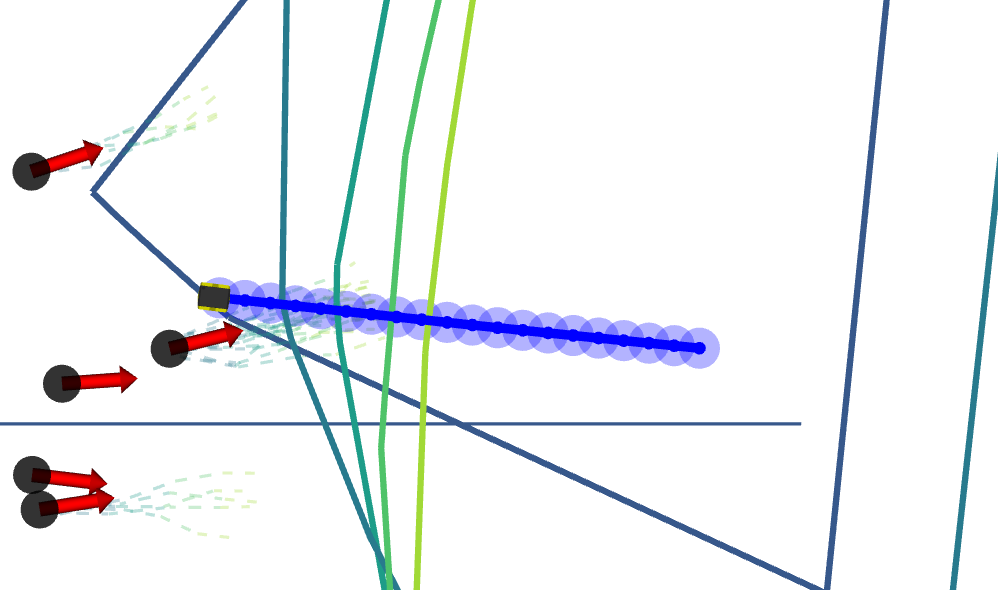}
  \caption{At $t=11.0$ s}
  \label{fig:sub2}
\end{subfigure}
\caption{Snapshots from the simulated environment under Gaussian pedestrian motion at different time instants. The pedestrians are represented by black circles where the red arrows indicate their direction of motion, blue circles depict the robot's planned trajectory. The constraints are visualized for stages 0, 5, 10, 15, and 19.}
\label{fig:simulation}
\end{figure*}

\begin{table*}[ht]
\caption{Statistical results over 100 experiments for a uni-modal simulation with 6 pedestrians. The comparison is done with respect to the maximum risk endured by the robot, duration, robot velocity, number of times the robot has to come to standstill, and minimum distance to the obstacles. The results are reported as ``average (standard deviation)''. The percentage of controller usage with $\epsilon = 0.2$, $\epsilon = 0.1$ and $\epsilon = 0.05$  is 87.98\%, 12.02\% and 0\%, respectively.}
\centering
\scalebox{0.95}{
\begin{tabular}{ ||c|c|c|c|c|c|c||}
 \hline
 \textbf{Upper Bound} & \textbf{Max CP} & \textbf{Dur. [s]} & \textbf{Vel. [m/s]} & \textbf{Temp. Freezing} & \textbf{Avg. Min Dist. [m]} & \textbf{Task Incomp.}\\
 \hline \hline
 $\epsilon=0.05$  & 0.0094 & 18.18 (0.169) &  1.15 (0.133) & 16 \% & 0.359 (0.06) & 13 \%\\
 \hline
  $\epsilon=0.1$  & 0.0456  & 17.24 (0.359) &  1.18 (0.186) & 15 \% & 0.308 (0.06) & 9 \%\\
  \hline
    $\epsilon=0.2$  & 0.0727  & 15.88 (0.317) &  1.24 (0.162) & 8 \% & 0.232 (0.06) & 4 \%\\
  \hline
  \hline
  \centering Hybrid $\epsilon =\{0.05, 0.1, 0.2\}$& \centering 0.0454 & 16.08 (0.176) & 1.23 (0.233) & 5 \% & 0.301 (0.07) & 6 \%\\
\hline
\end{tabular}}
\label{tableI}
\end{table*}

 \begin{table*}[ht]
\caption{Results similar to those in \ref{tableI} for uni-modal simulation with 10 pedestrians. The percentage of controller usage with $\epsilon = 0.2$, $\epsilon = 0.1$ and $\epsilon = 0.05$  is 87.45\%, 7.25\% and 5.30\%, respectively.}
\centering
\centering
\scalebox{0.99}{
\begin{tabular}{ ||c|c|c|c|c|c|c||}
 \hline
 \textbf{Upper Bound} & \textbf{Max CP} & \textbf{Dur. [s]} & \textbf{Vel. [m/s]} & \textbf{Temp. Freezing} & \textbf{Avg. Min Dist. [m]} & \textbf{Collision}\\
 \hline \hline
 $\epsilon=0.05$ & 0.0214  & 18.89 (0.324) &  1.01 (0.168) & 21 \% & 0.319 (0.07)& 0 \%\\
 \hline
  $\epsilon=0.1$  & 0.0636 & 18.04 (0.369) &  1.17 (0.141) & 18 \% & 0.264 (0.06) & 2 \%\\
  \hline
  $\epsilon=0.2$  & 0.1113  & 16.43 (0.302) &  1.20 (0.167) & 8 \% & 0.213 (0.05) & 3 \%\\
  \hline
  \hline
  \centering Hybrid $\epsilon =\{0.05, 0.1, 0.2\}$ & 0.0454  & 16.68 (0.185) & 1.19 (0.152) & 10 \% & 0.231 (0.09) & 0 \%\\ 
\hline
\end{tabular}}
\label{tableII}
\end{table*}
 
 \subsection{Simulation Results}
To create the reference path, a series of waypoints are defined and connected with a clothoid. The goal of the robot is to track the reference path as closely as possible while avoiding colliding with its surrounding dynamic obstacles, which are crossing freely. The baseline that we compare our results against is the scenario-based MPC approach proposed in \cite{SH-MPC}, with different risk levels $\epsilon$.
 In the following simulations, three scenario-based MPCs run in parallel with different acceptable risk levels, 0.05, 0.1, and 0.2. These values are chosen as a proof of concept of the proposed method. An upper bound of collision probability (CP), along a single planned trajectory, is set to 0.05. After each planning cycle, the maximum risk associated with each planned trajectory, from the three planners, is estimated according to \eqref{max_risk}, then the control commands from the controller with the least conservative solution, i.e., the one with the maximum risk level, are applied as long as the associated risk is less than $\epsilon_\circ = 0.05$. 
In case no feasible solution is obtained from all controllers without violating the threshold risk level, emergency braking is deployed so that the robot decelerates.\\ Several metrics are defined to compare the safety and efficiency of the proposed approach with the baseline. As safety metrics, we measure the maximum probability of collision per stage along the robot's planned trajectory, and the average minimum distance between the robot and the pedestrians, that is the distance between the robot's and pedestrian's circles' boundaries together with the percentage of physical collisions. As efficiency metrics, average speed, duration, and the temporary freezing percentage, that is the situations in which the robot has to come to a standstill in order to retain safety, are calculated. We consider a scenario as a temporary freezing scenario when the robot takes more than 2.0 s before it starts to accelerate again from a standstill. The reference velocity of the robot is set to 2.0 m/s whereas the velocity of the pedestrians is set to 1.0 m/s. The setup of the simulation is shown in Fig. \ref{fig:simulation}. 
  


 \subsubsection{Pedestrians with Gaussian noise} In the first scenario, the uncertainty of the pedestrian predictions is uni-modal Gaussian with a variance of $\boldsymbol{\Sigma}_w =0.5^2 \boldsymbol{I}$. We define the pedestrian dynamics as
 \begin{equation}
     \boldsymbol{\delta}_{k+1} = \boldsymbol{\delta}_k + (\boldsymbol{v} + \boldsymbol{\delta}_{w,k})dt, \quad \boldsymbol{\delta}_{w,k} \sim \mathcal{N}(\boldsymbol{0}, \boldsymbol{\Sigma}_w),
 \end{equation}
 
 \noindent where $v \in \mathbb{R}^2$ describes a constant velocity. Aggregated results in environments with 6 and 10 pedestrians, over 100 simulations, are presented in Tables \ref{tableI} and \ref{tableII} respectively. As shown in Table \ref{tableI}, the controller with $\epsilon = 0.05$ achieves the lowest collision probability compared to other controllers, however, at the expense of resulting in excessively  conservative trajectories. This conservatism can also be observed in the percentage of temporary freezing, in which the controller cannot find a solution that satisfies the risk bound along the planning horizon and the robot decelerates to a standstill. It can also be seen that the temporary freezing behavior decreases as the acceptable risk level increases, but this happens at the expense of violating the acceptable risk level $\epsilon_\circ$, for the controller with $\epsilon = 0.2$. On the contrary, by switching between the controllers based on the associated risk level, we can obtain less conservative results where the performance is comparable to the behavior of the controller with $\epsilon = 0.2$, but, most importantly, without violating $\epsilon_\circ$. Since the maximum collision probability for the controller with $\epsilon = 0.1$ never exceeds $\epsilon_\circ$, our method did not switch to the controller with $\epsilon = 0.05$, as the same safety level can be achieved with a less conservative behavior. A task is denoted as incomplete when the robot deviates from the reference path and does not get back to it by the end of the scenario while avoiding obstacles. The controller with $\epsilon = 0.2$, together with our method achieves the best performance with respect to task completeness. Similar behavior has been obtained in the environment with 10 pedestrians, as depicted in Table \ref{tableII}. In this environment the controller with $\epsilon = 0.2$ achieves the best performance with respect to the efficiency metrics, however it results in a higher maximum collision probability $11.1 \%$, and 3 physical crashes with one of the pedestrians. Again our approach manages to balance between safety and efficiency by obtaining shorter trajectories while providing a closer bound to the acceptable risk level, $0.0454/0.05$, and without leading to physical collisions, or many temporary freezing behavior.
 

 \begin{table*}[ht]
\caption{Statistical results over 100 experiments for a multi-modal simulation with 6 pedestrians. The comparison is done with respect to the maximum risk endured by the robot, duration, robot velocity, number of times the robot has to come to standstill, and minimum distance to the obstacles. The results are reported as ``average (standard deviation)''. The percentage of controller usage with $\epsilon = 0.2$, $\epsilon = 0.1$ and $\epsilon = 0.05$  is 83.67\%, 5.22\% and 1.11\%, respectively.}
\centering
\centering
\scalebox{0.95}{
\begin{tabular}{ ||c|c|c|c|c|c|c||}
 \hline
 \textbf{Upper Bound} & \textbf{Max CP} & \textbf{Dur. [s]} & \textbf{Vel. [m/s]} & \textbf{Temp. Freezing} & \textbf{Avg. Min Dist. [m]} & \textbf{Task Incomp.}\\
 \hline \hline
 $\epsilon=0.05$ & 0.0165   & 17.12 (0.717) &  1.22 (0.204) & 11 \% & 0.479 (0.16)& 9 \%\\
 \hline
  $\epsilon=0.1$  & 0.0693  & 15.88 (0.637) &  1.24 (0.185) & 6 \% & 0.407 (0.11) &  8 \%\\
  \hline
  $\epsilon=0.2$  & 0.1184  & 14.32 (0.673) &  1.29 (0.215) & 2 \% & 0.266 (0.12) & 6 \%\\
  \hline
  \hline
  \centering Hybrid $\epsilon =\{0.05, 0.1, 0.2\}$ & 0.0486  & 14.81 (0.655) & 1.29 (0.161) & 3 \% & 0.340 (0.13) & 1 \%\\ 
\hline
\end{tabular}}
\label{tableIII}
\end{table*}

\begin{table*}[ht]
\caption{Results similar to those in \ref{tableIII} for multi-modal simulation with 10 pedestrians. The percentage of controller usage with $\epsilon = 0.2$, $\epsilon = 0.1$ and $\epsilon = 0.05$  is 81.02\%, 15.22\% and 3.76\%, respectively.}
\centering
\scalebox{0.99}{
\begin{tabular}{ ||c|c|c|c|c|c|c||}
 \hline
 \textbf{Upper Bound} & \textbf{Max CP} & \textbf{Dur. [s]} & \textbf{Vel. [m/s]} & \textbf{Temp. Freezing} & \textbf{Avg. Min Dist. [m]} & \textbf{Collision}\\
 \hline \hline
 $\epsilon=0.05$ & 0.0234  & 18.34 (0.324) &  1.14 (0.258) & 13 \% & 0.426 (0.17)& 1 \%\\
 \hline
  $\epsilon=0.1$  & 0.0722  & 17.83 (0.369) &  1.19 (0.240) & 12 \% & 0.387 (0.16) & 6 \%\\
  \hline
  $\epsilon=0.2$  & 0.1396  & 15.89 (0.302) &  1.28 (0.223) & 8 \% & 0.291 (0.15) & 8 \%\\
  \hline
  \hline
  \centering Hybrid $\epsilon =\{0.05, 0.1, 0.2\}$ & 0.0452   & 16.12 (0.185) & 1.26 (0.292) & 6 \% & 0.329 (0.14) & 2 \%\\ 
\hline
\end{tabular}}
\label{table IV}
\end{table*}

\begin{figure*}
\centering
\begin{subfigure}{.24\textwidth}
  \centering
  \includegraphics[width=\linewidth]{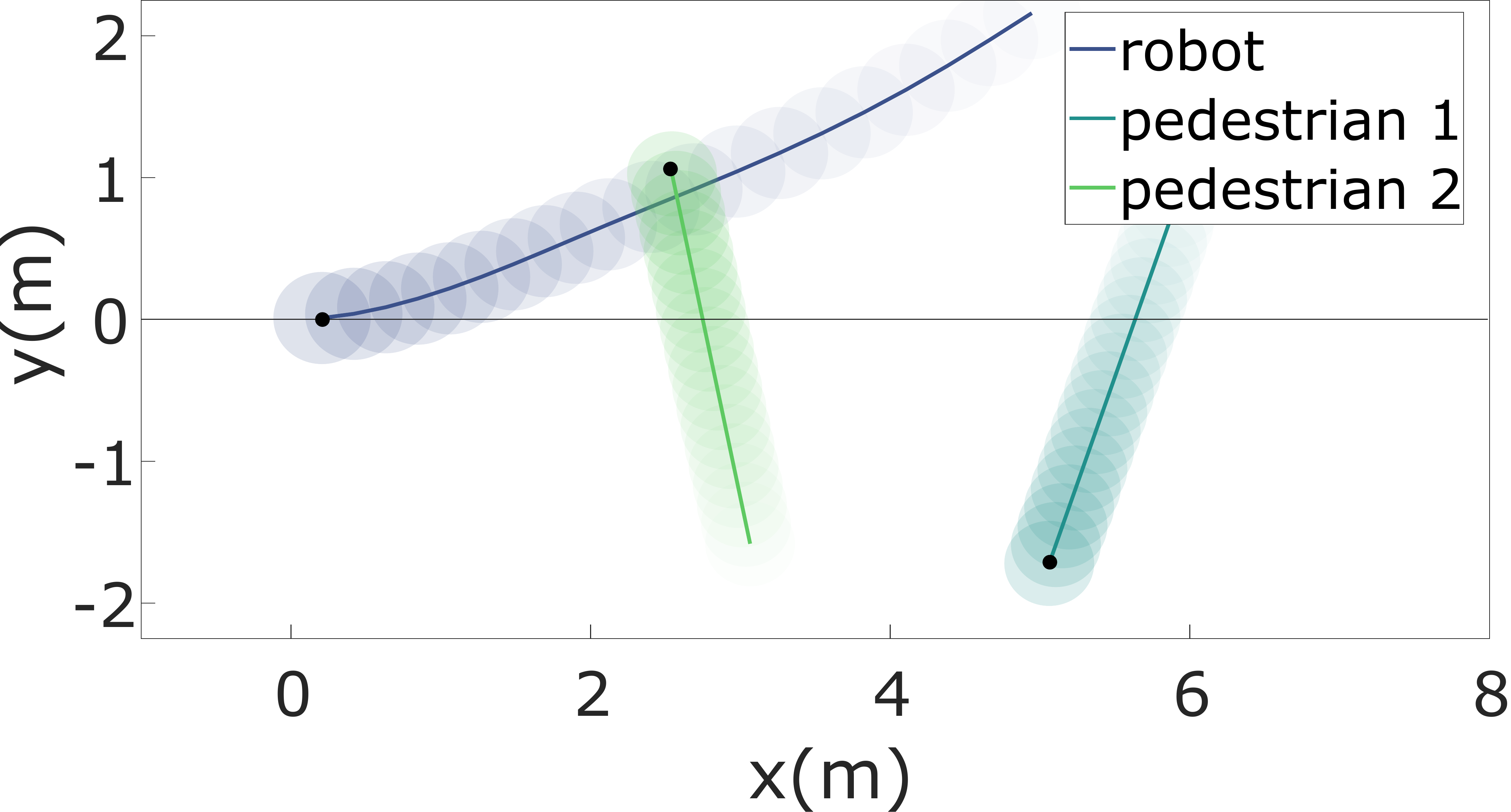}
  \caption{At $t=0.0$ s}
  \label{fig:sub1}
\end{subfigure}%
\begin{subfigure}{.24\textwidth}
  \centering
  \includegraphics[width=\linewidth]{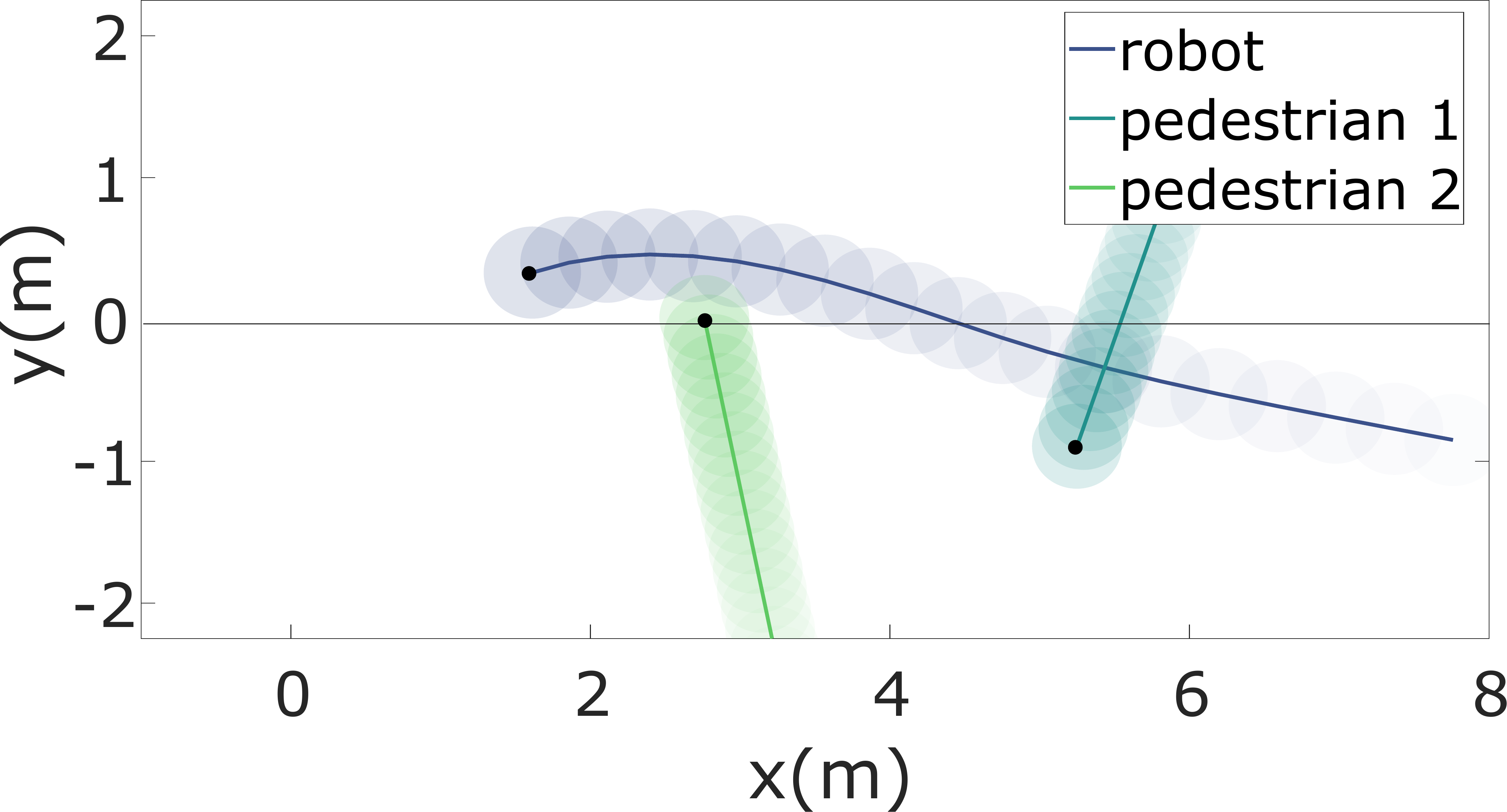}
  \caption{At $t=5.0$ s}
  \label{fig:sub2}
\end{subfigure}
\begin{subfigure}{.24\textwidth}
  \centering
  \includegraphics[width=\linewidth]{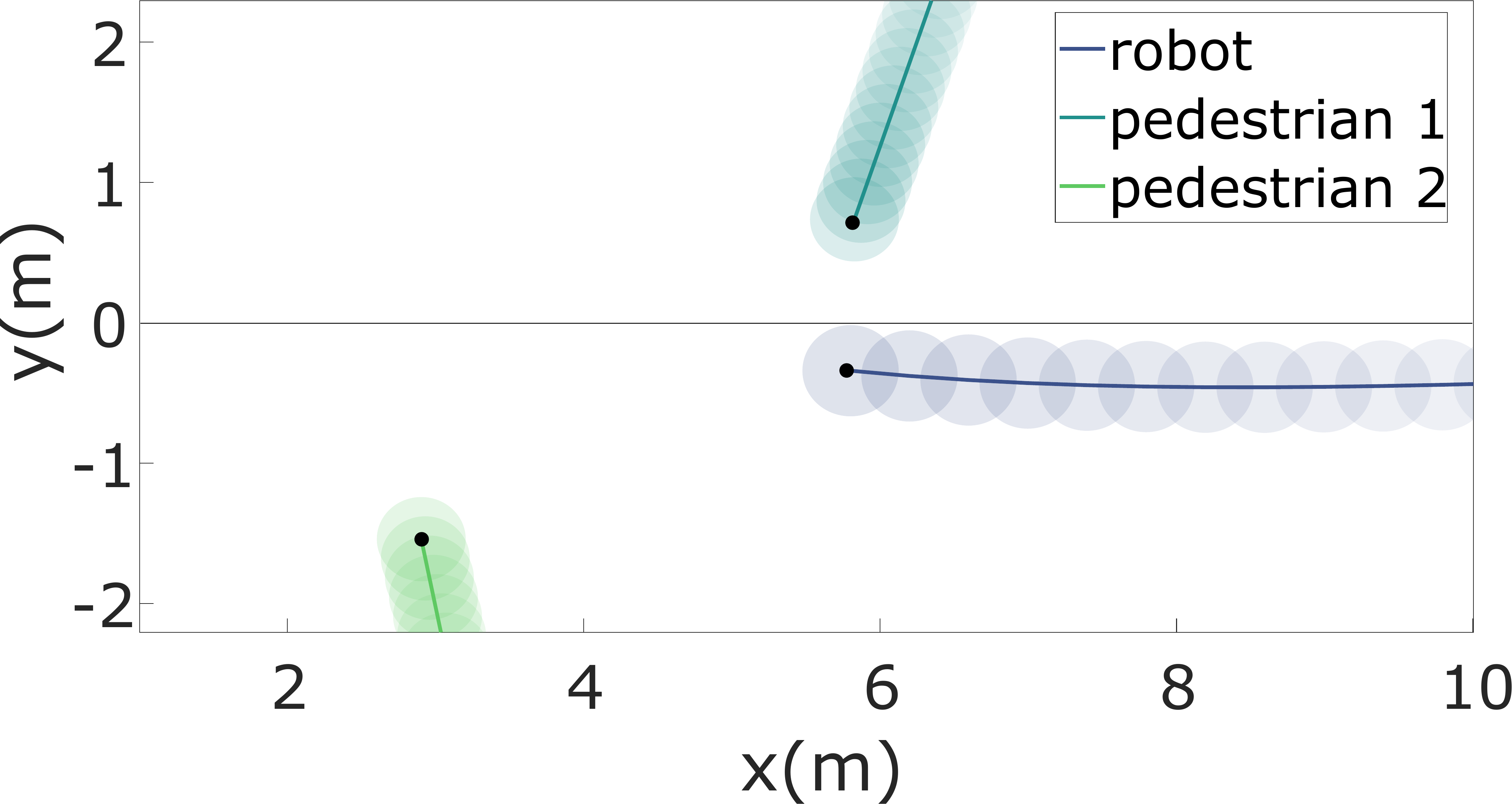}
  \caption{At $t=7.5$ s}
  \label{fig:sub2}
\end{subfigure}
\begin{subfigure}{.24\textwidth}
  \centering  \includegraphics[width=0.75\linewidth]{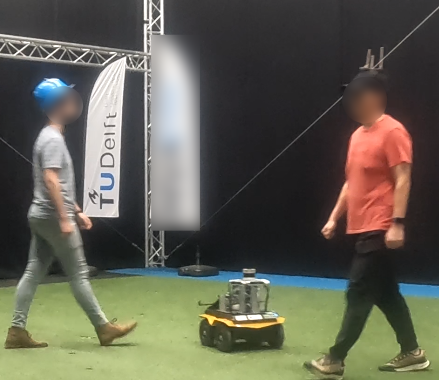}
  \caption{A snapshot from experiment}
  \label{fig:sub2}
\end{subfigure}
\caption{Experimental results with the robot avoiding two crossing pedestrians at different time instants. The blue circles depict the robot's plan, whereas the lime and green circles visualize the pedestrians' predictions where newer positions are depicted with lighter shades. The solid black line represents the reference path, and the black circles illustrate current positions. The robot takes 7.8 s to complete the task with 0.0416/0.05 max CP, 1.56 m/s average speed, and 25.784 ms computation time. The percentage of controller usage with $\epsilon = 0.2$, $\epsilon = 0.1$ and $\epsilon = 0.05$  is, 94.13\%, 5.87\% and 0\%, respectively.}
\label{fig:real}
\end{figure*}
\subsubsection{Pedestrians with Gaussian Mixture Model} In this section, we model the pedestrian movement by a Markov Chain that changes the pedestrian movement from horizontal to diagonal, with $p=0.975$ of staying in horizontal state and $p = 0.025$ of switching to diagonal state, in addition to the Gaussian noise of the previous simulation. The pedestrian dynamics are given by  
\begin{equation}
    \boldsymbol{\delta}_{k+1} = \boldsymbol{\delta}_k + (B\boldsymbol{v} + \boldsymbol{\delta}_{w,k}) dt, \quad \boldsymbol{\delta}_{w,k} \sim \mathcal{N}(\boldsymbol{0}, \boldsymbol{\Sigma}_w),
\end{equation}
\noindent where $B$ is either $B_h = \begin{bmatrix} 1 & 0
\end{bmatrix}^T$ or $B_d = \begin{bmatrix} \nicefrac{1}{\sqrt{2}} & \nicefrac{1}{\sqrt{2}} \end{bmatrix}^T$ based on the state of the Markov Chain. The uncertainties associated with this motion can be modeled as a Gaussian Mixture Model where each state transition in the Markov Chain leads to a separate mode with an associated probability (21 modes in total). Similar to the first scenario, the results are validated in environments with 6 and 10 pedestrians respectively. The results for 6 pedestrians are summarized in Table \ref{tableIII}. The results are in line with the 6-pedestrian Gaussian case. Here it can be noted that our approach outperforms the baselines on almost all risk metrics, while attaining, compared to the baseline with $\epsilon = 0.05$, a higher but still safe CP of 0.0486. By balancing the minimum distances to the pedestrians, the temporal freezing behavior is reduced compared to the other controllers which results in faster trajectories. Moreover, the robot manages to get back to the reference path in almost all simulations. For the case of 10 pedestrians, results are summarized in Table \ref{table IV}, where collisions occur for all methods in this environment. A significant improvement in the number of physical collisions, from 8$\%$ to 2$\%$, can be observed with respect to the baseline with $\epsilon = 0.2$ while attaining a comparable efficiency.  12.1$\%$ and 10.2$\%$ improvements are obtained in the trajectory duration, and speed respectively, compared to the baseline with $\epsilon = 0.05$. In all simulations, the average computation time of the full control loop of our approach is 74.68 ms with a maximum computation time of 91.16 ms which makes it real-time capable.

\subsection{Real-World Results}
We evaluated our method on a real robot navigating on a road, following the lane central line, while two pedestrians cross the road. A snapshot from our experiment\footnote{A video of the experiments and simulations accompanies this paper.} is shown in Fig. \ref{fig:real}, where quantitative results are illustrated in its caption. 


\section{Conclusion}
\label{conclusion}
In this paper, we showed that our proposed hybrid approach provides probabilistic safety guarantees while achieving a closer bound to the specified risk for a mobile robot operating among humans with uni-modal and multi-modal Gaussian uncertainties.  This is attained by running multiple probabilistic planners in parallel with different specified risk levels and quantifying the risk associated with each planned trajectory. 
The risk metric is estimated by integrating over the domain at which the robot trajectory and predicted pedestrian states spatially overlap at each stage along the prediction horizon. Accordingly, the plan with the least conservative behavior is chosen provided that its associated risk is below the risk level of the most conservative planner. Our simulations and experiments showed that the robot could follow the trajectories planned by the least conservative controller, 
most of the time, and only switches to more conservative controllers when the estimated risk violates the specified threshold. In such a way, the robot can plan faster trajectories while attaining the same safety level as the most conservative controller. 
Future works shall explore elaborated risk metrics in case the robot has a biased prediction for obstacle motion uncertainty and a criterion to set the risk upper bound for the probabilistic planner.  %


\bibliographystyle{IEEEtran}
\bibliography{IEEEabrv,IEEEexample}

\end{document}